\documentclass{IET}
\usepackage{graphicx}
\usepackage{caption}
\usepackage{refstyle}
\usepackage{amsmath}
\usepackage{url}
\usepackage{tikz} 
\usepackage{enumitem}
\usepackage{subfig}
\usepackage{multirow}
\usepackage{url}

\usepackage{titlesec}
\usepackage{fancyhdr}

\titleformat{\paragraph}
{\normalfont\normalsize\bfseries}{\theparagraph}{1em}{}
\titlespacing*{\paragraph}
{0pt}{3.25ex plus 1ex minus .2ex}{1.5ex plus .2ex}

\usepackage{makecell}
\usepackage{boldline}

\usepackage{array}  
\usepackage{dcolumn}

\usepackage[T1]{fontenc}

\usepackage{tabularx,ragged2e,booktabs,caption}
\usepackage{color, colortbl}
\usepackage{float}
\usepackage{array}

\usepackage [english]{babel}
\usepackage [autostyle, english = american]{csquotes}

\usepackage{tipa} 

\newcolumntype{C}[1]{>{\centering\arraybackslash}m{#1}}

\definecolor{Gray}{gray}{0.9}
\definecolor{LightCyan}{rgb}{0.88,1,1}
\definecolor{lightgray}{gray}{0.9}

\begin{document}

	\title{Experimental results on palmvein-based personal recognition by multi-snapshot fusion of textural features}
	
		\author[1]{\rm Mohanad Abukmeil}
	\author[2]{\rm Gian Luca Marcialis}

	\affil[1]{Department of Computer Science, University of Milan,  Via Giovanni Celoria, 18, 20133 Milano MI, Italy. E-mail: mohanad.abukmeil@unimi.it}
	
	\affil[2]{Department of Electrical and Electronic Engineering, University of Cagliari, Piazza d'Armi, I-09123 Cagliari, Italy. E-mail: marcialis@unica.it}
	
	\abstract{In this paper, we investigate multiple snapshot fusion of  textural features  for palmvein recognition including identification and verification. Although the literature proposed several approaches for palmvein recognition, the palmvein performance is still affected by identification and verification errors.  As well-known,  palmveins are usually described by line-based methods which enhance the vein flow. This is claimed to be unique from person to person. However, palmvein images are also characterized by texture that can be pointed out by textural features, which relies on recent and efficient hand crafted algorithms such as Local Binary Patterns, Local Phase Quantization, Local Tera Pattern, Local directional Pattern, and Binarized Statistical Image Features (LBP, LPQ, LTP, LDP and BSIF, respectively), among others. Finally, they can be easily managed at feature-level fusion, when more than one sample can be acquired for recognition. Therefore, multi-snapshot fusion can be adopted for exploiting these features complementarity. Our goal in this paper 	\footnotetext{The final version of this paper is in the production stage at Inderscience publisher, where it will be published in the International Journal of Biometrics (IJB) whin an ID (IJBM-240788).} is to show that this is confirmed for palmvein recognition, thus allowing to achieve very high recognition rates on a well-known benchmark data set.}
	
	\maketitle	

	\section{Introduction}\label{sec1}
	Biometric personal recognition systems are concerned with recognizing  individuals, or subjects on the other word, by either physiological traits   such as  face, iris, fingerprint, palmvein, and palmprint,  or by using some behavioral characteristics such as the voice or the  signature \cite{ISO,2,2 introduction1}.
	Among others, palmvein recognition is a promising technology which received considerable interest recently. The vein pattern residing in the palm have several characteristics  such as complex and rich structure, simultaneously and automatically assure the liveness in the presented sample, and difficult to imitate \cite{palmvein survay}. 
	
	Palmvein recognition uses the vascular patterns of an individual\textquoteright s palm as personal recognition data. Compared with a fingerprint \cite{2 introduction}, a palmvein has a broader and more complicated vascular pattern, and thus contains a wealth of differentiating features for personal recognition tasks. The palmvein is an ideal part of the body for this technology; it is not susceptible to visible occlusions or change in lighting conditions such as the facial features, or some environmental limitations due to the state of the skin (dryness) as fingerprint and  palmprint \cite{3 introduction, 2 section 2}.
	
	 Palmvein could be represented by some lines and texture features, which are  constituted from veins. Different types of palmvein features could be extracted according to the application type where they are used \cite{2 introduction,7 introduction}. Palmveins are often represented by the vein flow of the hand which is claimed to be unique from person to person. Palmvein verification  could be  considered for several daily applications including on-line banking, ATMs, access control, time attendance management, and E-commerce. Its main use has been so far in palmvein verification, for example, checking the football team fans identity\footnote{\url{http://news.trust.org/item/20141010100845-b881i?view=print}}, but, as we will show in this paper beside verification, palmvein identification should be considered for immigrants checking and forensics investigation as well. In fact, palmvein can't be faked or canceled as fingerprints (immigrants tend to hide their fingerprints in order to impede the possibility to be identified) \cite{palmvein survay}.

	Single-modal palmvein, single-sample palmvein recognition has been intensively studied \cite{1 introduction,6 introduction}. However, two points should be taken into account: among other approaches, the ones relying on textural features have not yet fully been taken into account, whilst in our and others opinion \cite{6 introduction 1,ltp,LBP section 2} the palmvein could be described in terms of textural sets to gain better accuracy. In particular, many hand crafted textural features extraction approaches have been developed recently, and some of them exhibited the general ability to extract significant information for the task at hand, independently of the biometric trait adopted \cite{6 introduction 8}. It is worth to point  that the methods used for feature extraction and combination are not novel and used in some recent works as in \cite{7 introduction}, nevertheless, the novelty of our work is in the comprehensive experiments that show the complementary information inherent to multiple instances and patterns of the palmvein.
	
	
	
	In this context, the contributions of this paper are threefold: (1) we test extensively a subset of recent hand crafted  local descriptor, namely, Local Binary Patterns (LBP \cite{LBP}),  Local Phase Quantization (LPQ \cite{LPQ}), Local Tera Pattern (LTP \cite{ltp}), Local Directional Pattern which didn't used yet for palmvein  (LDP \cite{ldp}), and Binarized Statistical Image Features (BSIF \cite{BSIF}) in both verification and identification use-cases; (2) we  empirically  investigated at which extent the multi-snapshot fusion of palmvein images, at feature level, can further improve the verification and identification performance;
	(3) we explored a statistical driven analysis based on recommended performance evaluation indicators released by ISO \cite{ISO} and \cite{indicators} to reach comparable and fair reproducible research. The recommended indicators are: (\textit{i}) Equal Error Rate \textit{EER} which presents the single index measure when FAR and FRR intersect. (\textit{ii}) Half total error rate \textit{HTER} is estimated by averaging FAR and FRR, using this indicator will fix the cost function when rejecting genuine subject or accept another imposter. (\textit{iii}) Genuine Acceptance Rate \textit{GAR} which defines the percentage of accepted subject at the point of EER. For the identification evaluation we use (\textit{iv}) \textit{Rank-1} which outcome is considered as correct identification.

	The task at hand here is palmvein recognition including identification and verification. In identification,  the experiments are done on closed set, where the identity of the subject must be found on a set of possible candidates (watch-list). Therefore, no information about the subject's identity is given. In verification, the subject  makes a claim as to its identity, and the goal is to determine the authenticity of the claim. In this case, the probs, that is, the palmvein images submitted during sytem's operations are compared only with the stored subject templates (biometric reference) of the claimed identity.
	Experiments are carried out on a publicly available benchmark data set. We show that (1) among tested algorithms, BSIF outperformed existing approaches, (2) feature-level fusion may further improve the performance by allowing to achieve very high recognition rates.
	
	
	The rest of this paper is organized as follows: Section 2 reports textural representation of palmvein, Section 3  discusses feature extraction and fusion. Section 4 reports the experimental results and the performance evaluation of our experiments. Section 5 concludes the paper.

	\section{Textural representation of a palmvein}
	\subsection{Palmveins as set of texture}
	\label{Local dense descriptor and feature extraction}
	
\begin{figure*}[hbt!]
		
		\centering
		\includegraphics[scale=.6] {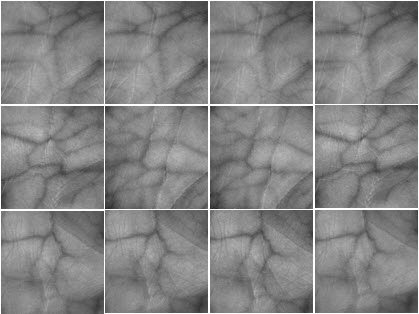}	
		
		\caption {\textit{Vein texture pattern  of human palm. When palmvein image is acquired, differently of palmprint image, only the vein  pattern is visible as  dark lines (and curves) texture, therefore the recognizer must translate the texture pattern and match it with the previously stored patterns of the subject in the biometric reference database. The palmvein images  in this figure are qouted from PolyU Hyper-Spectral Database$^2$  }.} 
		
		\label{fig:palmveintexture}
		
	\end{figure*}

	The texture  can be considered as a similarity gathering of sub-patterns in a palmvein image \cite{section 2}. The local sub-pattern characteristics of palmvein give rise to the perceived uniformity, roughness, lightness, and density. Palmvein trait is described by arterial and venous ramifications \cite{1 section 2}. The characteristic shapes of vein features are similar to principal lines and wrinkles features of palmprint \cite{3 section 2,3 section 22},  thus there is a great potential to exploit the information from palmvein sub-pattern (texture entities) for recognition tasks. Fig.~\ref{fig:palmveintexture} has been taken from the PolyU Hyper-Spectral database\footnote{\url{http://www4.comp.polyu.edu.hk/biometrics/HyperspectralPalmprint/HSP.html}} and shows ROI of palmvein and the blood vein texture pattern.
	
	Textural descriptors have a broad spectrum of applications and can describe palmvein images as well. In particular, a certain Region of Interest (ROI) is extracted and  divided into a set of patches, therefore a set of filters is applied to each patch; the filter responses are collected in terms of a bit set to 1 according to the positive correlation between patch and filter, set to 0 otherwise \cite{LBP,BSIF,LPQ}. In this way the n-dimensional bit string for each ROI is defined; finally, the frequency of each bit string configuration is computed and the feature set is composed by a set of frequencies.  It is evident that, independently of the textural descriptor, the \textit{i}-th features have the same meaning, because it is directly related to the same bit string configuration with respect to the adopted filter.

	

	In the following section we describe the investigated textural descriptors. They are also called patch-based image features \cite{local dense descriptor section3}, among others, Local Binary Pattern (LBP), Local Phase Quantization (LPQ), Local Tera Pattern (LTP), Local Directional Pattern (LDP), and Binarized Statistical Image Features (BSIF). These techniques have become increasingly popular also in biometric recognition applications \cite{6 introduction 1}. In this section, we give a brief introduction about the proposed texture algorithms.
	The simplicity and robustness of such descriptors (local dense) make it dominant among other feature extraction techniques. 
	
	\subsection{Local texture descriptors} 
	
	In this section, we will consider five local dense descriptors: LBP, LPQ, LTP, LDP, and BSIF, which are used to extract the textural feature from  plamvein, and for sake of space, more details are given in each technique's reference.
	
	 {\textit{LBP} is a rotation invariant and gray-scale dense descriptor \cite{LBP}. For each target pixel, $x$, it considers $P$ uniformly sampled neighbors on square or circle (called batch or window) of radius $R$ centered on $x$, when the position of neighbor does not coincide with a pixel site resorting to interpolation is required. According to the linear filters and  threshold, the gray value of $x$ is compared  with all other neighbor pixels.  If the value of $x$ is greater or lower  than any values of its neighbor, then the result will be \textit1 (in the case of lower), or \textit0 (in the case of greater).  Therefore, in each patch, the result is a  vector (of each palmvein image) containing binary code, also called bit string, of length equal to $P$. LBP has been  proposed in the literature of palmvein recognition in \cite{LBP section 2} and \cite{2 LBP section 2}.  
		
		\textit{LPQ} descriptor is similar to LBP in adopting a binary encoding scheme \cite{local dense descriptor section3}. To make the descriptor  insensitive to centrally symmetric blur, the patch surrounding the target pixel $x$ is analyzed in the Fourier domain. Over each patch, the related phase information is extracted by Short-Time Fourier Transform (STFT) leading to extract four complex-valued coefficients at each selected frequency, and binary quantization is introduced to get a form of 8-bit feature. 
		
		\textit{LTP} is considered also extended version of LBP, but it analyses the batch in  form of  three values, which are $1$, $0$, and $-1$. LTP is considered less noise sensitive, as it encodes the pixel into separated states. The descriptor quantize based on the patch of width {$\pm t$} and gray center pixel $g_c$. Three status are considered as quantization results; the pixel values are quantize to $-1$ for those below $(g_c-t)$, quantize to $+1$ for those above $(g_c+t)$, and those around $g_c$ are quantized to $0$. Additional details about LTP can be found in \cite{ltp}.   
		
		\textit{LDP} by using this descriptor, the spatial structure of local features could be characterized. The binary code is computed by considering the value of each pixel location in all 8 directions. LDP encodes the palmvein texture by using edge responses values of 8  neighboring pixels, differently from LBP which rely on the intensity changes around the pixels. LDP operator  extracts several features such as junctions, corner, and curves which are constitute vein structure. Given the central pixel $g_C$, the eight possible directional edge response values ${m_i}, i=0,1,2,..,7$ are computed in 8 orientations by using Kirsch masks $M_i$. More details about LDP are summarized in \cite{ldp}.

		\textit{BSIF} is inspired from LPQ and LBP techniques \cite{BSIF}. In both techniques  (LBP and LPQ) each pixel's neighborhood   is analyzed in binary encoding scheme to get a binary code for each palmvein image, whereas in BSIF a set of filters is used for image convolution and the filter responses are binarizing in 8-bit code string. BSIF depends on a set of automatically predefined filters, thus it operates in contrast of LPQ and LBP which based on manually predefined filter. A set  13 of natural images is used to learn the filters, similarly to other local dense descriptor, the bit string is generated by binarizing the response of a linear filter (with threshold set to zero). The length of code string indicates how many filters are used, and each bit in the string is associated  with a different filter.
		
		\section{Feature extraction and proposed fusion scheme}
		\subsection{Feature extraction}
		\label{sec:Feature extraction}
		In this paper, we used LBP, LPQ, LTP, LDP and BSIF as feature extractors. Such techniques compute a binary code string for the pixels of a given  ROI of palmvein. The code value of a pixel is considered as a local descriptor of the image intensity pattern in the pixel's surroundings. Further, histograms of pixel's code values taken into account; it allows to characterize the texture properties within ROI, also  of a particular configuration of the filters outputs. Specifically, the recent textural techniques have been applied so far to several biometric traits (for example, fingerprint, fingervein, and face), always exhibiting very high recognition rates \cite{BSIF for fingerprint,BSIF for face,ldp, face bsif}. Therefore, we investigated if its application field can be extended to improve the palmvein identification and verification rates based on the proposed  protocol by Raghavendra \textit{et al.} in \cite{ragavendar}.

		The verification system scheme we investigated in this paper is depicted in Fig.~\ref{fig:verificationscheme1}. Where, in the case of identification, a classification model is required after fusion stage to identify the subjects. It is worth noting that all textural analysis techniques adopted in this work are common in the number of bit	 string (n=8). 8-bit stranded coding results feature vector with 256 coefficients, which presented by a histogram is used in our work,  see Fig.~\ref{fig:verificationscheme1}. The filters used in BSIF are learnt  from a set of 13 natural images by using Independent Component Analysis according to \cite{BSIF}. We will report if the used filters in BSIF can improve the recognition performance when training by palmvein images instead of natural images, as a sub-goal. Further details will be given in Section \ref{Experimental results}.

		\begin{figure*}[hbt!]
			
			\begin{center}
				\includegraphics  [scale=.5] {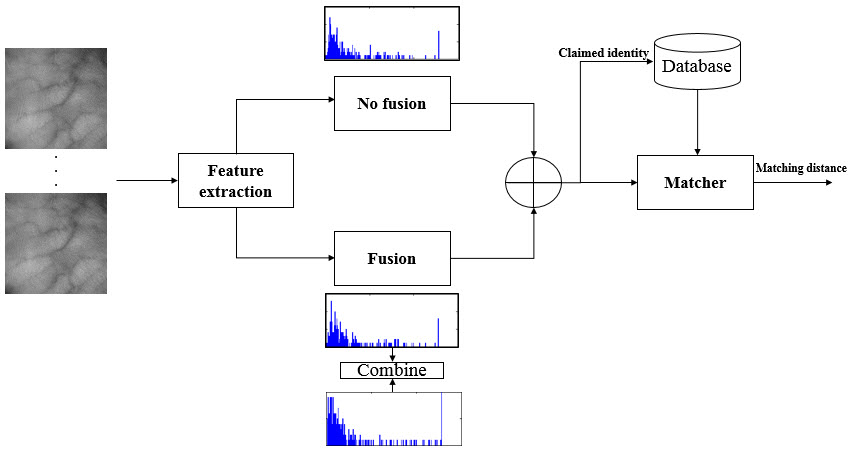}	
				
				\caption {\textit{Block diagram of the proposed palmvein verification. Once ROI is obtained, the features are extracted by the proposed textural methods and characterized by a histogram for each ROI. Where, the feature vector of the claimed identity comes from no fusion, or multi-snapshot module and compared with the previously stored one in the database, using a matcher (for verification task) that is replaced by classifier the case of identification.}}
				
				\label{fig:verificationscheme1}
				
			\end{center}
			\vspace{-10pt}	
		\end{figure*}
		
		\subsection{Multi-snapshot fusion}
		
		It is well-known by large sets of experimental evidences \cite{8 introduction, handbookfusion} that multi-biometrics systems, that is, systems able to exploit the complementarity coming from multiple biometric sources, can greatly improve both verification and identification performance. According to the literature,  potentiality of multi-snapshot fusion have not yet fully investigated for palmvein recognition. Among the three fusion levels, namely, feature, match score, and decision level \cite{handbookfusion}, the first one is done before obtaining the similarity (match score) between two models of palmvein images, while, score level fusion is done after finding out match score. Feature and score levels fusion were compared in \cite{feature level}, whereas the best results  obtained by relying feature level fusion.
		
		The current literature exhibits prevalently works when palmvein images are processed by one or more than textural descriptors, combined by a score-level approach, which requires the independent computation of scores. For example, Ref.\cite{LBP section 2} adopted LBP and LDP (Local derivative pattern) score-level fusion, but experimental results have not been satisfying. Al-juboori \textit{et al.} \cite{motivation} proposed the fusion of global features obtained by wavelet coefficients computation, and LBP to obtain local feature where the performance improvement was obtained . Yan \textit{et al.} show that multi-snapshot fusion is worthy to be applied in \cite{7 introduction}, where they used Scale Invariant Feature Transform (SIFT) to extract palmvein features, However, this work didn't mentioned which combination rule the authors adopted for fusion. Raghavendra \textit {et al.} \cite{ragavendar} proposed a novel fusion scheme, at score level, of palmvein images based on Discrete wavelet Transform (DWT). However, the results was sufficient in their, but different of our works since the performance evaluated at score level fusion and depending on CASIA multispectal palmprint database (only 100 subjects). Our proposed work will be evaluated at feature level fusion depending on large database (380 subjects). Palmvein biometrics have been also combined with palmprint\cite{8 introduction,fusion3} and signature\cite{fusion2}, but \textit{multi-modal} fusion is out of this paper scope. 
		
		The proposed approach implies that more than one palmvein image of the same subject are available in enrollment and recognition stages. For each subject, there are several images stored in biometric reference database. In our proposed work, the features of each image are extracted  using textural descriptors, and can be characterized by a histogram which represents the frequency responses of textural filters output. 

		The choice of using multiple snapshots instead of other approaches like the multi-algorithmic ones (which would require only one sample per user) is motivated by the nature of the adopted features, due to the strong characterization of each textural descriptor. In fact a multi-algorithmic fusion approach, namely, the computation of a fusion rule to combine the outputs of different descriptors is not appropriate, in our opinion.  Moreover, each of the adopted textural descriptor computes a feature vector which corresponds to the histogram of the frequencies of a particular configuration of the filters outputs. In other words, since each filter represents a peculiar textural template, how frequently of a certain template can be detected with other ones is estimated. This is equivalent to evaluate the frequency of patterns, whose locality is referred to the size of the patch (see the previous Section). 
		
      Factually,   the estimation of feature vector or pattern  can be affected by errors \cite{6 introduction 1}. Therefore, we averaged such estimation by taking into account more samples of the same image and a given textural descriptor, to reduce such estimation errors. This agrees with the theory that the sample mean standard deviation is less than that of the individual estimation. Moreover, such mean tends to the true mean of the related random variable under investigation (\textit{i.e.}, the true frequency of the local textural configurations). It is reasonable to hypothesize that the obtained averaged features set is more reliable and representative of the subject identity than the individual features set obtained by processing only one image. This should be true for both training and probe sets. Consequently, during systems operations (the probing stage), the Euclidean distance between the enrolled template and the probe would be lower if the sample comes from a genuine user, and higher otherwise, thanks to the averaging process. At the same time, this effect should be of great help in the identification stage, by making closer features sets coming from the same subjects and farer features sets coming from different subject, thus leading to a better CMC curve.
		
		The averaging choice is connected to the choice of using the Euclidean distance as the matching score estimator, which is inversely proportional to such distance. The Euclidean distance is defined as $d(x,y)=\sum{d_i^2}=\sum{(x_i-y_i)^2}$. If $x_i$ is the mean of all $y_i, \forall y=\{y_1, ...,y_h\}$ ($h$ is the feature space size), then  $d$ is minimized because each $d_i$ is minimized. This is true from the geometrical point of view. From the statistical point of view, $x$ is the sample mean from a set of measurements $y^{(1)},...,y^{(m)}$, that is, our template set of a given subject. For a genuine sample $z$, coming from the same distribution of $y$, $d(x,z)$ should be still near to the minimum, that is, corresponding to the sum of the minimum variances from the individual features; whilst for impostors samples, which are not coming from the same distribution of $y$, $d$ should be even greater, on average, than all possible distances $d(y,z), \forall y$. This is even true if the average operation is performed in both the enrollment and probing stages, because in this case  averaging samples of the same impostor or of the same genuine user, thus making them farer or closer to the enrolled template than the individual probes. 
		
		
		On the basis of motivations above, for each subject $i$, multiple feature vectors are stored in the biometric reference data set. For instance, the subject $i$ in the data set have $(S_{(i,1)},S_{(i,2)},..,S_{(i,n)})$ feature vectors. The simple averaging rule we used as a follow, in order to obtain the fused template set:
		
		\begin{equation}		
		S_{i_{fused}}=
		\{S_{(i_{fused},1)}, S_{(i_{fused},2)},...,S_{(i_{fused},{n \choose 2})}\}=
		\{\dfrac{S_{(i,1)}+S_{(i,2)}}{2}, \frac{S_{(i,2)}+S_{(i,3)}}{2},..., \frac{S_{(i,n-1)}+S_{(i,n)}}{2}\}
		\end{equation}
		
		where, $S_{i_{fused}}$ are the  fused feature sets of the first subject, and $S_{(i,n)}$ represents the last feature vector, and so on for the reminder subjects, see  Fig.~\ref{fig:fusion}. 
		Accordingly, the following fusion steps have been taken into account:
		
		\begin{itemize}
			\item Features from the same subject are extracted form their ROIs by using the aforementioned methods. 
			\item For each subject $i$ a set of templates are obtained as $S_i$: $S_i=\{S_{i,1},S_{i,2},...,S_{i,n}\}$.
			\item The extracted feature vectors are normalized and centered at zero mean to remove the redundancy, that is, each vector is such that the sum of features is 1. 
			\item The extracted feature vectors are combined (\textit{i.e.} fused in pairs) to derive $S_{i_{fused}}$ as shown in Eq. (1), thus obtaining ${n \choose 2}$ templates.
		\end{itemize}		
		
		We also explored the use-case in which the fusion is also performed on the probe samples, that is, a couple of samples is also required during the system's operations. 
		The obtained averaged feature set is more representative and reliable of the subject identity than individual feature set (\textit{i.e. no fusion}) \cite{6 introduction 1}. The euclidean distance is selected as biometric matcher which described in \cite{duda book} and the final match score is inversely proportional to such distance. In order to give a full view of our system, let us consider, for example that, during the system operation, the subject $i$ is claimed.The subject is eventually asked to submit two samples, namely $X_1$ and $X_2$. They are averaged in order to obtain $X$. 
		The euclidean distance between the templates and the probe $X$ is given by:
		\begin{equation}
		d(S_{i_{fused}},X)=\sqrt{\sum_t{(S_{(i,n)}-X)^2}}
		\end{equation}
		
		The fusion, in our proposed work, is carried out in pairs, meaning that, every two vectors are fused together to exploit the complementarity features, see Fig.~\ref{fig:fusion}. As written above, $X$ may be also obtained by the fusion of a pair of probe samples in the same manner adopted for the templates. It could be hypothesized that combining more than two feature vectors could lead to an accuracy even better. However, we did not investigate this case because we must take into account, especially during system's operations, the trade-off between the performance and the user's cooperation level \cite{7 introduction}. The methodology  with more than two samples during testing stage may improve the performance at the expense of reducing the acceptance and the ease-of-use of the whole system \cite{7 introduction}.    
			\begin{figure*}[hbt!]
			
			\begin{center}
				\includegraphics  [scale=.55] {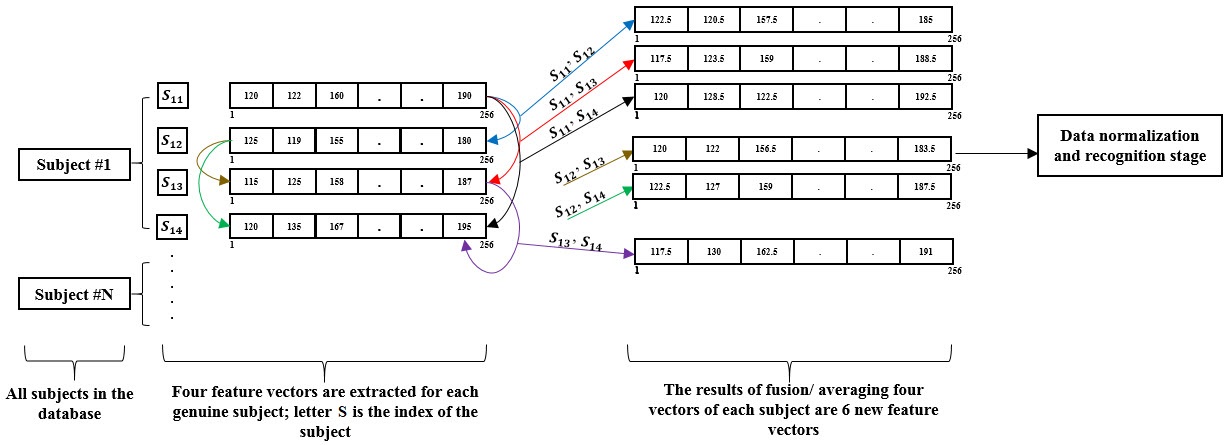}	
				
				\caption {\textit{Fusion of $n=4$ templates for genuine subject $S$, where $S_{1,1}$ means that the first feature vector of the first subject in the database, and so on. For each subject $S$, basic templates were fused in pairs by averaging, thus obtaining ${n \choose 2}$ templates, moreover the same for probs. }}
				
				\label{fig:fusion}
				
			\end{center}
			\vspace{-10pt}	
		\end{figure*}  
		
		The literature proposes different works in the context of multi-snapshot fusion \cite{6 introduction 1,motivation} (multi-sampling in some works), but the lack of inclusiveness in fusion rules, texture feature descriptors, and statistical significance of some works  has stimulated us to give deep investigation to fill this gap. In this paper, the multi-snapshot fusion is extensively studied; the proposed combination procedure is simple and theoretically well-founded; finally, we show by experiments that both identification and verification accuracies are strongly improved.

		\section{Experimental results and discussion}
		\label{Experimental results}

		This section describes the used data set, performance evaluation protocol, and the experimental results of our proposed work for identification and verification modes. 
		
		\begin{itemize}
			\item In identification mode, the histogram of each  given palmvein from the probe set is extracted and compared with all histograms stored in the template set to determine which subject it is most similar to. The \textit{k}- nearest neighbor search with euclidean distance metric is applied to determine identity among all enrolled subjects. 
			
			\item In verification mode, the extracted histogram  is compared with all histograms of the \textit{claimed identity} stored in the template set.
		\end{itemize}
		
		The overall recognition system performance is assessed statistically, depending on confidence interval,  using Rank-1 indicator for identification. Furthermore,  EER, GAR at the point of EER, and HTER  which generated by varying the thresholds on  euclidean distance  measure for verification. aforementioned indicators are described in Section \ref{sec1}. 
		
		\subsection{Data Set description}
		The performance of our proposed work has been evaluated by using the PolyU Hyper-Spectral palmprint data set. A novel sensing device has been developed by the Biometric Research Center at  the Hong Kong polytechnic university, which can acquire images from palmprint  with different bands (wavelength) \cite{PolyU2}. The wavelength range from 420$nm$ to 1100$nm$, and has been used to build a large-scale hyper-spectral palmprint data set, and the age distribution starts from 20 to 60 years. For each wavelength, both palms were involved and 7 images taken from each one. On overall, the database comprises 5,240 images from 380 different palms for each  wavelength. The data sets  were used also in \cite{PolyU1,table2 3}. According to \cite{PolyU2}, the most discriminant features could be extracted when using 0960, 0860, and 0700$nm$ as sensor wavelengths, thus our experimental results comprises the performance of three different data sets as a function of wavelength. 
		It is worthy to mention that, when we got the data set and prepared it for experiments, we found that for each subject, there are only 12 valid images captured for each palm in both sessions. Accordingly, our experimental works employ all subject in the database and fixing the number (which is 12) of images for each palmvein.
		
		Table\ref{tab1}  gives some details of the used data set. The size of data set is varied to investigate the effectiveness of our proposed approach, from the smallest number of templates, namely, two, and a maximum of four templates to maintain the  probs set size big enough. For each  Subject, twelve palmvein images are considered and subdivided in two non-overlapped sets, randomly chosen. The first subset is the template set (also called biometric reference according to ISO terminology \cite{ISO}), the second one is the probe.

		\begin{table*}[hbt!]
			\resizebox{0.99\textwidth}{!}{
				\centering \arraybackslash
				
				\begin{tabular}{@{}|c|c|c|c|c|c|c|c|c|c|c|c|c|@{}} \hline \hline
					
					\cellcolor {LightCyan}	\textbf {Protocol}  & \multicolumn{5}{c|}{No fusion} & \multicolumn{5}{c|}{Multi-snapshot fusion} \\
					\hline
					\rowcolor{lightgray}
					Texture feature & LTP & LDP & LQP  & LBP & BSIF & LTP & LDP & LQP  & LBP & BSIF \\
					
					\hline
					\# Templates &  \multicolumn{5}{c}{ 2 \hspace{1cm} 3 \hspace{1cm} 4} \vline &   \multicolumn{5}{c}{ 1 \hspace{1cm} 3 \hspace{1cm} 6}  \vline \\
					\hline
					\rowcolor{lightgray}
					\# Probs &  \multicolumn{5}{c}{ 10 \hspace{0.9cm} 9 \hspace{0.9cm} 8} \vline &   \multicolumn{5}{c}{ 45 \hspace{0.8cm} 36 \hspace{0.8cm} 28}  \vline \\
					\hline
					
					Pixel dimension  & \multicolumn{9}{c}{128 x 128 pixel}  & \\
					\hline
					\rowcolor{lightgray}
					Print dimension  & \multicolumn{9}{c}{1.33 x 1.33 inch}  & \\
					\hline
					Bitdepth  & \multicolumn{9}{c}{8 bit "gray level"}  & \\
					\hline
					\rowcolor{lightgray}
					Resolution  & \multicolumn{9}{c}{96 "dbi"} & \\
					\hline
					
					Used bands  & \multicolumn{9}{c}{0960, 0860, and 0700$nm$ } & \\
					\hline

					
					

			\end{tabular}}	
			
			\caption{Data set description. This table reports the number of templates and probes   for individual algorithms including no fusion  and  multi-snapshot fusion scenarios. The data set  size is varied by considering two, three and four templates per subject, therefore ten, nine, and eight probes per subject (respectively), to show the recognition rate variation as function of the biometric reference data set size. Other image information details of the data set are included as well.}
			
			\label{tab1}
		\end{table*}

		The adopted local dense descriptors  offered publicly by The Center for Machine Vision Research (CMV)\footnote{\url{http://www.cse.oulu.fi/CMV/Downloads}}, and  have been evaluated in Matlab environment.
		
		\subsection{Evaluation protocol}
		\label{Evaluation protocol}
		Our proposed work is evaluated based on raghavendra \textit{et al.}\cite{ragavendar} testing protocols, that measure the performance of Hyper-Spectral palmprint data set. We tested three data sets based on it's band ,namely, 0700, 0860, and 0960$nm$. The following is a description of both protocols:
		
		\textbf{Protocol 1:} This protocol is designed to evaluate performance of palmvein biometric based on our proposed scheme, considering several variations such as pose, time, noise, etc, in the used data sets. Each PolyU data set partitioned into two unequal subsets, namely, probs set and template set. The partition  process was randomly repeated for $n$ times (where $n$ = 10), without overlapping and using holdout cross-validation \cite{ragavendar,protocol}. For each subject, we selected 4 ROI images as template set and 8 ROI images as probs set; later the size of both set will be varied for the sake of comparison. Consequently, for each trial (\textit{i.e.} experiment in the other word), in no fusion case,  we have $380 \times 4 = 1520$ genuine comparison scores, while the imposter comparison are equal to $380 \times 379 \times  8 = 1152160$  scores. The experiments are done on all ten  different partitions, and the mean of all runs (10 runs) is taken  to present the results. Moreover,  the $90\%$ parametric confidence interval reports the statistical variations of our experiments \cite{protocol}.
		
		\textbf{Protocol 2:} To investigate  the reliability of our proposed scheme, it must to evaluate the performance according to time variation effect. Accordingly, we partitioned each data set into two equal subsets ,namely, templates set which taken from the first session and probs set taken from the second session of the PolyU data sets. Consequently, for each trial, in no fusion case, we have $380 \times 6 = 2280$ genuine comparison scores, whereas the imposter scores are  $380 \times 379 \times  6 = 864120$. The experiments are done on all two runs, one run by considering the first session as template set and the other by switch it to be probs sets, also the mean of all runs taken to present the results.

		Among all proposed feature extraction methods which suggested in Section \ref{sec:Feature extraction}, BSIF (which is much focused in our work) requires tuning and fixing some parameters such as filters length, patch size, and training according to what reported in \cite{BSIF}. Thus, we built a development set to fix the wanted parameters by selecting 100 random  subjects from the data set, also in multi snapshot fusion such set is required to fix the best combination rule. For instance, we noticed that, for BSIF, the best filter length is 8-bit, the superior patch size $17 \times 17$ pixel, and no change maintained in performance when filters training based on natural  or palmvein images. For multi-snapshot fusion, we found that averaging the features in pair gives the best result. All parameters values are fixed utilizing the development set and kept constant over the performance evaluation using the aforementioned protocols. Finally, our results are reported according to standard terminology \cite{ISO}, and in the terms of EER, GAR at EER, HTER, and Rank-1 indicators. More description about these indicators is given in Section \ref{sec1}.
	 
	\subsection{Results and discussion}
	
	The first set of our experiments carried out  based on baseline biometric system (\textit{i.e no fusion}). In the first stage we used the development set to fix the experiments parameters, then the performance  evaluation of no fusion scenario and the proposed multi snapshot combination is reported.	
			
				\subsubsection{Individual algorithms performance (no fusion)}
				\label{no fusion}
			The baseline biometric  performance  is evaluated in this scenario. Samples coming from probe sets were  classified to its corresponding classes (identification), or compared in other task with previously stored templates of claimed identity (verification). 
			
			According to the literature, among the textural descriptors  BSIF  haven't been fully investigated for palmvein recognition. In particular, the BSIF-based method is characterized by the choice of different filters and patches size \cite{BSIF}. We performed several experiments in the context of verification, and utilizing the development data set to fix the best configuration of BSIF filters and patch size. ROC curve is used to present the performance, where the best ROC obtained from  $17 \times 17$ patch size with $8-bit$ filter length as shown in Fig.~\ref{fig:BSIF_Patch_size}, thus the best patch size and filter length are fixed in all other experiments of BSIF.
				
				\begin{figure}[ht!]
					\begin{center}
						
						\includegraphics[width=0.40\linewidth]{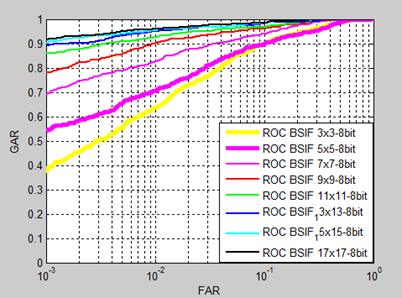}
					\end{center}
					\caption{ Roc curves of BSIF as a function of window size, where 4 templates and 8 probs were considered for each subject. The best ROC curve is obtained from 17 x 17 window size with 8-bit filter length.}
					
					\label{fig:BSIF_Patch_size}
					
				\end{figure}

				Because BSIF exploits filters learnt from natural images by Independent Component Analysis \cite{BSIF}, we also attempted to train a novel set of filters by using a set of palmvein images, but we noticed no change in the performance. Adopting natural images for training the filters gave better recognition rates than using palmvein images. We hypothesized that using natural images avoided the problems of data set variance and over-fitting, thus allowing to derive a general set of filters, namely, textures which may be characterized whichever image. Actually, this is an interesting issue which could deserve further and deeper investigations beyond the topic reported here. Fig.\ref{fig:BSIF_training}  illustrates the $ROC$ curves to show the  difference between both BSIF training.

				\begin{figure}[hbt!]
					\begin{center}
						
						\includegraphics[width=0.40\linewidth]{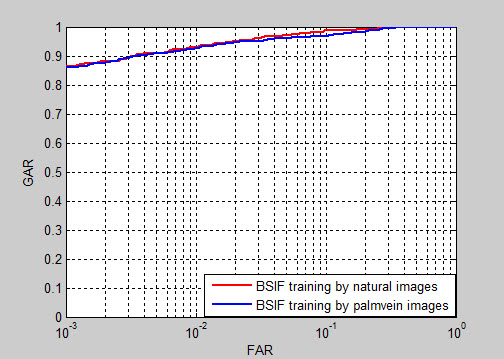}
					\end{center}
					\caption{ The ROC curve without fusion comprising FAR and GAR to show the difference between natural and palmvein images training of BSIF filters. For each subject 4 templates and 8 probs were considered, where,  the red and blue curves represent ROC curves by training  BSIF filters based on natural images and palmvein images, respectively. Also, the big difference can be noticed around $FAR=10^{-1}$.}
					\label{fig:BSIF_training}
					
				\end{figure}
			\paragraph {Result on protocol-1}
			
			\label{Result on protocol-1}
				
			During baseline biometric evaluation,  we considered the both protocols which aforementioned in Section~\ref{Evaluation protocol}. Two scenarios are considered in the first protocol evaluation; the first scenario built based on 4 templates and 8 probs for each subject. Accordingly, the first scenario comprises $380 \times 4 = 1520$ genuine scores and $380 \times 379 \times 8 = 115216$ imposer scores. Table.\ref{no fusion first protocol first scenario} below shows the four performance indicators which adopted in our work, where the best result is taken from BSIF and the less abundant results is given from LTP. Moreover, the best data set band is obtained when the wavelength is equal to $0960nm$, which confirms what \cite{PolyU2} claimed on choosing  suitable spectral band for palmvein recognition.  The indicators for BSIF, and $960nm$ data set gave results as $EER=3.23 \pm 1.01\%$, $ GAR @ EER= 96.77 \pm 1.01\%$, $minHTER=2.83 \pm 0.91\%$, and $Rank-1= 94.61 \pm 2.21\%$.

			\begin{table*}[hbt!]
				\centering
				\resizebox{0.8\textwidth}{!}{
					
					\begin{tabular}{@{}|c|c|c|c|c|c|@{}} \hline \hline
						
						\multicolumn{5}{|c|}{ \cellcolor {LightCyan} $\mathbf{First\, protocol-first \,scenario  \ ( 4 \  templates \ \  \& \ \ 8\ probs )}$} \\
						\hline
						
						\multirow{2}{*} {$\mathbf{Method}$} & \multirow{2}{*} {$\mathbf{Indicator}$} & \multicolumn{3}{c|} {$\mathbf{Data \, set\, band}$} \\
						
						\cline{3-5} & & {$\mathbf{0700\textit{nm}}$} & {$\mathbf{0860\textit{nm}}$} & {$\mathbf{0960\textit{nm}}$} \\
						\hline

						\multirow{4}{*}{$\mathbf{LTP}$} &  {${EER}$} &$26.14\pm1.22$& $18.99\pm1.49$&$19.02\pm1.48$\\ \cline{2-5} 
						& 	${GAR @ EER}$ & $73.88\pm 1.23$ & $81.01\pm1.50$ & $80.97\pm1.48$ \\ \cline{2-5}
						
						&{${minHTER}$} & $25.75\pm1.11$ & $18.44\pm1.29$ & $18.80\pm1.39$ \\ \cline{2-5}
						&	{${Rank-1}$} & $29.78 \pm2.01 $ & $ 11.29 \pm1.07$ & $44.82\pm2.79$\\
					\hline
					
						\multirow{4}{*}{$\mathbf{LDP}$} &  {$EER$} & $12.24\pm1.16$ & $11.29\pm1.07$ & $11.00\pm1.26$ \\ \cline{2-5} 
						& 	{${GAR @ EER}$} &$ 87.76\pm1.16$ &$ 88.70\pm1.06 $& $89.00\pm1.26$ \\ \cline{2-5}
						&{	${minHTER}$} & $12.10\pm1.12$ &$ 11.02\pm0.99$ & $10.90\pm1.25$ \\ \cline{2-5}
						&	{${Rank-1}$} & $42.63\pm3.66$ & $50.99\pm3.96 $& $44.52\pm3.20$\\
					\hline

					\multirow{4}{*}{$\mathbf{LPQ}$} &  ${EER}$ &$ 12.05\pm1.13 $& $18.37\pm6.02$ & $10.14\pm1.54$  \\ \cline{2-5} 
					& 	{${GAR @ EER}$} &$ 87.96\pm1.13$ &$ 81.63\pm6.02$ & $89.88\pm1.55$ \\ \cline{2-5}
					&	{${minHTER}$} & $11.80\pm1.13$ &$ 17.95\pm6.09$ & $9.94\pm1.48$ \\ \cline{2-5}
					&	{${Rank-1}$} & $54.77\pm4.17$ &$ 46.75\pm9.77$ &$ 61.80\pm4.61$\\
					\hline
						
					\multirow{4}{*}{$\mathbf{LBP}$} &  {$EER$} & $10.81\pm1.20$ & $11.62\pm1.18$ & $9.82\pm1.30$ \\ \cline{2-5} 
					& 	{${GAR @ EER}$} & $89.18\pm1.20$ & $88.38\pm1.18 $& $90.18\pm1.29$ \\ \cline{2-5}
					&	{${minHTER}$} & $10.66\pm1.17 $& $11.33\pm1.19$ & $9.74\pm1.30 $\\ \cline{2-5}
					&	{${Rank-1}$} & $57.88\pm4.61$ & $61.62\pm5.19$ & $64.14\pm4.83$\\
				\hline
						
						\multirow{4}{*}{$\mathbf{BSIF}$} &  {$EER$} & $6.71\pm1.19$ & $7.92\pm1.94$ & $\mathbf{3.23\pm1.01}$ \\ \cline{2-5} 
						& ${GAR @ EER}$& $93.29\pm1.19$ & $92.08\pm1.95 $& $\mathbf{96.77\pm1.01}$ \\ \cline{2-5}
						&	${minHTER}$ & $6.37\pm1.17$ & $7.00\pm1.86$ &  $\mathbf{2.83\pm0.91 }$\\ \cline{2-5}
						&	${Rank-1}$ & $86.84\pm3.30$ & $85.05\pm5.33$ & $\mathbf{94.61\pm2.21}$\\
						\hline
						
				\end{tabular}}
				
				\caption{Verification and identification performance of no fusion.  The first scenario of the first protocol with $90\%$ confidence interval is considered, where, Method is the feature extraction technique; Indicator is biometric performance parameter given in \%; Data set band is the wavelength used to collect each data set. }
				
				\label{no fusion first protocol first scenario}
				
				\end{table*}

		The second scenario of the first protocol tests the robustness of feature extraction methods when the size of templates and probs are changing. We considered two cases of data size variations; inside the first case we decreased the size of templates (for each subject) to be 3 templates and 9 probs, where the genuine scores become $380 \times 3 = 1140$ and the imposter comparison  are $380 \times 379 \times 9 = 1296180$ scores. In the second case, we reduced the templates to be 2 templates with new genuine comparison scores $380 \times 2 = 760$,  also the probs increased to be 10 and the imposter comparison scores are equal to  $380 \times 379 \times 10 = 1440200$. We considered $EER$ as a biometric performance indicator.
		 On comparing the results of both cases, we noticed that among all feature extraction methods BSIF is outperformed all other, even when changing the size. Also, the best results of both cases are given  when using $0960nm$ data set, BSIF feature extractor, and 3templates with $EER$ equal to $5.05 \pm 0.07\%$. Accordingly, we noticed that varying both templates and probs size is important to find which feature extraction method gives the best results, also to investigate the reliability of the proposed work. 4 templates and 8 probs, 3 templates and 9 probs, and 2 templates and 10 probs are the sizes variations with associated  $EER$ for BSIF: $3.23 \pm 1.01\%$, $ 3.41 \pm 0.81\%$, and $5.05 \pm 0.07\%$, respectively, see Table.\ref{no fusion first protocol first scenario} and Table.\ref{no fusion first protocol second scenario}.

		 	\begin{table*}[hbt!]
		 	\centering
		 	\resizebox{0.8\textwidth}{!}{
		 		
		 		\begin{tabular}{@{}|c|c|c|c|c|c|c|c|@{}} \hline \hline
		 			
		 			\multicolumn{8}{|c|}{ \cellcolor {LightCyan} \textbf {First protocol- second scenario}} \\
		 			\hline
		 			
		 			\multirow{2}{*} {\textbf{Method}} & \multirow{2}{*} {\textbf{Indicator}} & \multicolumn{3}{c|} {\textbf{3 templates \& 9 probs}} & \multicolumn{3}{c|} {\textbf{2 templates \& 10 probs}}\\
		 			
		 			\cline{3-8} & & {\textbf{0700\textit{nm}}} & {\textbf {0860\textit{nm}}} & {\textbf{0960\textit{nm}}} & {\textbf{0700\textit{nm}}} & {\textbf {0860\textit{nm}}} & {\textbf{0960\textit{nm}}}  \\
		 			\hline
		 			\textbf{LTP} & EER & $26.47 \pm 0.74$ & $19.31 \pm 0.98$ & $19.12 \pm 1.05$ & $28.40 \pm 0.41$ & $21.51 \pm 0.62$ & $20.93 \pm 0.20$ \\
                     \hline
                     
                     \textbf{LDP} & EER & $12.39 \pm 0.90$ & $11.67 \pm 0.71$ & $11.35 \pm 0.93$ & $14.55 \pm 0.15$ & $13.39 \pm 0.30$ & $13.22 \pm 0.13$ \\
                     \hline

                     \textbf{LPQ} & EER & $12.54 \pm0.86$ & $18.88 \pm 5.84$ & $10.47 \pm 1.13$ & $14.43 \pm 0.12$ & $21.46 \pm 5.24$ & $12.73 \pm 0.18$ \\
                     \hline

                     \textbf{LBP} & EER & $11.22 \pm 0.93$ & $12.09 \pm 0.91$ & $10.09 \pm 1.02$ & $13.34 \pm 0.17$ & $14.31 \pm 0.43$ & $12.23 \pm 0.16$ \\
                     \hline
                     \textbf{BSIF} & EER & $7.30 \pm 0.89$ & $8.61 \pm 1.51$ & $\mathbf{3.41\pm0.81}$ & $5.05 \pm 0.07$ & $12.05 \pm 0.56$ & $\mathbf{5.05 \pm 0.07}$ \\
                     \hline

		 	\end{tabular}}
	 
 \caption{EER  performance indicator of no fusion,  for the second scenario of the first protocol with $90\%$ confidence interval, where, Method is the feature extraction technique; Indicator is biometric performance parameter (EER) given in \%; Data set band is the wavelength used to collect each data set.}

\label{no fusion first protocol second scenario}

\end{table*}

\paragraph{Result on protocol-2}
\label{Result on protocol-2}
The ability to recognize the subjects considering time variation effect is measured adopting this protocol, more details are given in Section \ref{Evaluation protocol}. Only one scenario is evaluated, which measures the performance of the first session with respect th the second session, and vise versa.  The  genuine comparison scores are equal to $380 \times 6 = 2280$, and $380 \times 379 \times 6 = 864120$ . As it can be seen from Table.~\ref{no fusion second protocol}, in the context of identification, LPQ is the worst method for feature extraction where the most subjects are not identified; LPQ gives the lower $Rank-1$ on comparing with all other methods. The verification performance of BSIF is outperformed all others techniques and LPQ is put in the bottom of the table, where, $EER$ of $0960nm$ data set are $9.32 \pm 0.28\%$ and $48.64 \pm 0.224\%$, respectively.

	\begin{table*}[hbt!]
		\centering
		\resizebox{0.8\textwidth}{!}{
			
			\begin{tabular}{@{}|c|c|c|c|c|c|@{}} \hline \hline
				
				\multicolumn{5}{|c|}{ \cellcolor {LightCyan} $\mathbf{Second\, protocol \ ( 6 \  templates \ \  \& \ \ 6\ probs )}$} \\
				\hline
				
				\multirow{2}{*} {$\mathbf{Method}$} & \multirow{2}{*} {$\mathbf{Indicator}$} & \multicolumn{3}{c|} {$\mathbf{Data \, set\, band}$} \\
				
				\cline{3-5} & & {$\mathbf{0700\textit{nm}}$} & {$\mathbf{0860\textit{nm}}$} & {$\mathbf{0960\textit{nm}}$} \\
				\hline

				\multirow{4}{*}{$\mathbf{LTP}$} &  {${EER}$} &$31.02\pm0.19$& $31.18\pm10.34$&$24.98\pm0.47$\\ \cline{2-5} 
				& 	${GAR @ EER}$ & $69.01\pm0.18$ & $68.82\pm10.32$ & $75.04\pm0.51$ \\ \cline{2-5}
				
				&{${minHTER}$} & $30.54\pm0.08$ & $30.13\pm10.50$ & $24.79\pm0.21$ \\ \cline{2-5}
				&	{${Rank-1}$} & $18.55\pm1.30 $ & $21.01\pm14.36 $ & $29.43\pm0.07$\\
				\hline
				
				\multirow{4}{*}{$\mathbf{LDP}$} &  {$EER$} & $16.35\pm0.27$ & $18.60\pm6.35$ & $15.56\pm0.35$ \\ \cline{2-5} 
				& 	{${GAR @ EER}$} &$83.64\pm0.29 $ &$ 81.38\pm6.39 $& $84.45\pm0.32$ \\ \cline{2-5}
				&{	${minHTER}$} & $16.19\pm0.48$ &$18.34 \pm6.19$ & $15.42\pm0.43$ \\ \cline{2-5}
				&	{${Rank-1}$} & $27.02\pm2.53$ & $24.39\pm17.03$& $30.37\pm0.69$\\
				\hline

				\multirow{4}{*}{$\mathbf{LPQ}$} &  ${EER}$ &$ 50.36\pm0.18 $& $50.29\pm0.27$ & $51.35\pm0.28$  \\ \cline{2-5} 
				& 	{${GAR @ EER}$} &$49.63\pm0.18 $ &$ 49.69\pm0.29$ & $48.64\pm0.22$ \\ \cline{2-5}
				&	{${minHTER}$} & $49.81\pm0.03$ &$49.79\pm0.22 $ & $49.21\pm0.03$ \\ \cline{2-5}
				&	{${Rank-1}$} & $0.11\pm0.18$ &$0.15\pm0.04 $ &$ 0.13\pm0.22$\\
				\hline
				
				\multirow{4}{*}{$\mathbf{LBP}$} &  {$EER$} & $15.07\pm0.02$ & $17.48\pm3.37$ & $14.39\pm0.63$ \\ \cline{2-5} 
				& 	{${GAR @ EER}$} & $84.93\pm0.04$ & $ 82.50\pm3.39$& $85.61\pm0.58$ \\ \cline{2-5}
				&	{${minHTER}$} & $14.82\pm0.04 $& $17.24\pm3.45$ & $14.19\pm0.50 $\\ \cline{2-5}
				&	{${Rank-1}$} & $41.03\pm3.28$ & $33.86\pm9.09$ & $45.33\pm1.41$\\
				\hline
				
				\multirow{4}{*}{$\mathbf{BSIF}$} &  {$EER$} & $10.30\pm0.27$ &  $19.14\pm9.95$& $\mathbf{9.32\pm0.28}$ \\ \cline{2-5} 
			& ${GAR @ EER}$&  $89.71\pm0.25$ & $80.88\pm9.96$ &  $\mathbf{93.68\pm0.29 }$ \\ \cline{2-5}
			&	${minHTER}$ & $9.88\pm0.37$ & $18.60\pm10.05$ &  $\mathbf{ 5.74\pm0.27}$\\ \cline{2-5}
			&	${Rank-1}$ & $75.77\pm0.11$ & $41.93\pm37.88$ & $\mathbf{87.28\pm0.72}$\\
			\hline
				
		\end{tabular}}
		
		\caption{Verification and identification performance of no fusion.  The second protocol with $90\%$ confidence interval is considered, where, Method is the feature extraction technique; Indicator is biometric performance parameter given in \%; Data set band is the wavelength used to collect each data set.}
		
		\label{no fusion second protocol}
		
	\end{table*}
\pagebreak
	
\subsubsection{Multi-snapshot fusion}
\label{Multi snapshot fusion}	
For each subject, $S$ basic templates were fused in pairs by averaging, thus obtaining ${n \choose 2}$ templates. During system's operations (recognition phase), two images are acquired and related feature vectors fused with  each other. The resulting feature vector is compared with previously obtained templates stored in the biometric reference database. In this work, we attempted to test several fusion rules utilizing development set as done by Mirmohamadsadeghi \textit{et al.} \cite{6 introduction 1}, to adopt the best of them in multi-snapshot fusion of palmvein. We concluded that the simple "average" fusion rule  is the best rule on the basis of our experiments setup, accordingly  average rule (mean) is fixed in all fusion experiments. Fig.~\ref{fig:fusion rule} shows the ROC curves based on several fusion rules and BSIF features, which exhibits the best results in baseline palmvein biometric, see Section.\ref{no fusion}. The following are the results of the proposed fusion scheme based on Raghavendra's \textit{et al.} evaluation protocols which described in Section.\ref{Evaluation protocol}.  

	\begin{figure}[hbt!]
	\begin{center}
		
		\includegraphics[width=0.40\linewidth]{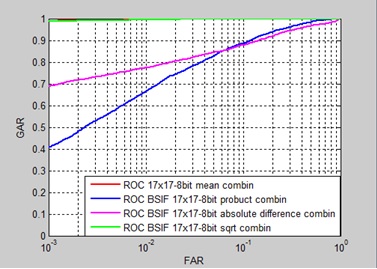}
	\end{center}
	\caption{ BSIF ROC curve of multi-snapshot fusion comprising FAR and GAR to show the difference between several fusion rules based on BSIF features. Red, green, blue, and magenta curves represent the ROC when using  mean, sqrt, product, and absolute difference rules, respectively.}
	\label{fig:fusion rule}

\end{figure}
	
\paragraph{Results on protocol-1}Two scenarios are considered for performance evaluation as in Section ~\ref{Result on protocol-1}. For each subject, the number of templates increased from $4$ to $6$ also probs from $8$ to $28$, thanks to fusion process. Inside the first scenario, we compared $380 \times 6 = 2280$ genuine scores and $ 380\times379\times28=4032560$ imposter scores. Our score comparison is considered very large comparing with other existing works to meet the reliability of our fusion scheme. As it can be concluded from Table.~\ref{fusion first protocol first scenario}, BSIF outperform all other feature extraction methods, furthermore, data set with wavelength band $0960nm$ gives the best results with $EER= 1.74 \pm 0.87\%$, $GAR @ EER= 98.26 \pm 0.88\%$, $ HTER= 1.50 \pm 0.76\%$ and $Rank-1= 97.62 \pm 1.43\%$.

\begin{table*}[hbt!]
	\centering
	\resizebox{0.8\textwidth}{!}{
		
		\begin{tabular}{@{}|c|c|c|c|c|c|@{}} \hline \hline
			
			\multicolumn{5}{|c|}{ \cellcolor {LightCyan} $\mathbf{First\, protocol-first \,scenario \ ( 6 \  templates \ \  \& \ \ 28\ probs ) }$} \\
			\hline
			
			\multirow{2}{*} {$\mathbf{Method}$} & \multirow{2}{*} {$\mathbf{Indicator}$} & \multicolumn{3}{c|} {$\mathbf{Data \, set\, band}$} \\
			
			\cline{3-5} & & {$\mathbf{0700\textit{nm}}$} & {$\mathbf{0860\textit{nm}}$} & {$\mathbf{0960\textit{nm}}$} \\
			\hline

			\multirow{4}{*}{$\mathbf{LTP}$} &  {${EER}$} &$19.34 \pm 2.18$& $13.57\pm2.16$&$13.05\pm2.41$\\ \cline{2-5} 
			& 	${GAR @ EER}$ & $80.66\pm2.18$ & $86.44\pm2.16$ & $86.96\pm2.42$ \\ \cline{2-5}
			
			&{${minHTER}$} & $19.06\pm2.06$ & $13.14\pm2.06$ & $12.83\pm2.33$ \\ \cline{2-5}
			&	{${Rank-1}$} & $43.11\pm4.87 $ & $ 59.23\pm6.78$ & $62.53\pm7.56$\\
			\hline
			
			\multirow{4}{*}{$\mathbf{LDP}$} &  {$EER$} & $7.51\pm1.71$ & $7.72\pm1.35$ & $7.05\pm1.71$ \\ \cline{2-5} 
			& 	{${GAR @ EER}$} &$ 92.48\pm48$ &$92.74\pm1.35  $& $92.95\pm1.70$ \\ \cline{2-5}
			&{	${minHTER}$} & $7.47\pm1.69$ &$7.14\pm1.31 $ & $6.97\pm1.67$ \\ \cline{2-5}
			&	{${Rank-1}$} & $61.96\pm7.11$ & $67.88\pm6.52 $& $65.73\pm7.19$\\
			\hline

			\multirow{4}{*}{$\mathbf{LPQ}$} &  ${EER}$ &$ 7.17\pm1.52 $& $14.43\pm6.85$ & $6.53\pm1.92$  \\ \cline{2-5} 
			& 	{${GAR @ EER}$} &$92.83\pm1.52$ &$85.59\pm6.81$ & $93.47\pm1.92$ \\ \cline{2-5}
			&	{${minHTER}$} & $7.04\pm1.49$ &$14.17\pm6.87 $ & $6.31\pm1.85$ \\ \cline{2-5}
			&	{${Rank-1}$} & $72.58\pm6.60$ &$59.50\pm13.45$ &$79.59\pm7.44 $\\
			\hline
			
			\multirow{4}{*}{$\mathbf{LBP}$} &  {$EER$} & $6.22\pm1.16$ & $7.12\pm1.48$ & $5.56\pm1.67$ \\ \cline{2-5} 
		& 	{${GAR @ EER}$} & $93.78\pm1.61$ & $92.88\pm1.47$& $94.44\pm1.66$ \\ \cline{2-5}
		&	{${minHTER}$} & $ 6.13\pm1.57$& $7.00\pm1.46$ & $5.51\pm1.64 $\\ \cline{2-5}
		&	{${Rank-1}$} & $76.99\pm6.70$ & $76.62\pm7.18$ & $82.69\pm6.93$\\
		\hline
			
			\multirow{4}{*}{$\mathbf{BSIF}$} &  {$EER$} & $3.42\pm1.23$ & $4.77\pm1.95$ & $\mathbf{1.74\pm0.87}$ \\ \cline{2-5} 
		& ${GAR @ EER}$& $96.58\pm1.23$ & $95.23\pm1.95$& $\mathbf{98.26\pm0.88}$ \\ \cline{2-5}
		&	${minHTER}$ & $3.20\pm1.18$ & $4.43\pm1.83$ &  $\mathbf{ 1.50\pm0.76}$\\ \cline{2-5}
		&	${Rank-1}$ & $94.66\pm2.63$ & $90.95\pm4.36$ & $\mathbf{97.62\pm1.43}$\\
		\hline
			
	\end{tabular}}
	
	\caption{Verification and identification performance of multi snapshot fusion.  The first scenario of the first protocol with $90\%$ confidence interval is considered, where, Method is the feature extraction technique; Indicator is biometric performance parameter given in \%; Data set band is the wavelength used to collect each data set.  }
	
	\label{fusion first protocol first scenario}

\end{table*}

The second scenario of the first protocol includes two cases of data set size variation as a result of multi- snapshot fusion. The first case comprises   $3$ templates and $36$ probs, while in the second case the templates decreased to $2$ and the probs increased to $ 45$ probs. Accordingly, $380 \times 3 = 1140 $ genuine scores and $380 \times 379 \times 36 = 5184720 $ imposter scores  considered for comparison in  the first case, moreover  $380 \times 2= 760 $ genuine scores and $380 \times 379 \times 45 = 6480900 $ imposter scores taken into account for comparison in the second case. Table~\ref{fusion first protocol second scenario} comprises the performance evaluation of both cases, where the best $EER= 2.93 \pm 0.06\%$ taken from $BSIF$ feature extraction method and data set wavelength band equal to $0960 nm$, as well, the least efficient performance taken from $LTP=23.56 \pm 0.41\%$ and $0700nm$ data set band.

	\begin{table*}[hbt!]
	\centering
	\resizebox{0.8\textwidth}{!}{
		
		\begin{tabular}{@{}|c|c|c|c|c|c|c|c|@{}} \hline \hline
			
			\multicolumn{8}{|c|}{ \cellcolor {LightCyan} \textbf {First protocol- second scenario}} \\
			\hline
			
			\multirow{2}{*} {\textbf{Method}} & \multirow{2}{*} {\textbf{Indicator}} & \multicolumn{3}{c|} {\textbf{3 templates \& 36 probs}} & \multicolumn{3}{c|} {\textbf{1 templates \& 45 probs}}\\
			
			\cline{3-8} & & {\textbf{0700\textit{nm}}} & {\textbf {0860\textit{nm}}} & {\textbf{0960\textit{nm}}} & {\textbf{0700\textit{nm}}} & {\textbf {0860\textit{nm}}} & {\textbf{0960\textit{nm}}}  \\
			\hline
			\textbf{LTP} & EER & $19.94\pm1.52$ & $14.04\pm1.48$ & $13.19\pm1.77$ & $23.56\pm0.41$ & $17.18\pm0.35$ & $16.54\pm0.21$ \\
			\hline
			
			\textbf{LDP} & EER & $7.83\pm1.25$ & $7.48\pm1.00$ & $7.22\pm1.26$ & $10.55\pm0.13$ & $9.36\pm0.19$ & $9.52\pm0.10$ \\
			\hline

			\textbf{LPQ} & EER & $7.48\pm1.12$ & $14.86\pm6.59$ & $6.59\pm1.41$ & $10.38\pm0.15$ & $17.47\pm5.96$ & $9.24\pm0.15$ \\
			\hline

			\textbf{LBP} & EER & $6.46\pm1.17$ & $7.40\pm1.03$ & $5.69\pm1.25$ & $9.22\pm0.14$ & $9.82\pm0.28$ & $8.03\pm0.09$ \\
			\hline
			\textbf{BSIF} & EER & $3.70\pm0.85$ & $5.13\pm1.48$ & $\mathbf{1.74\pm0.70}$ & $5.74\pm0.11$ & $7.80\pm0.22$ & $\mathbf{2.93\pm0.06}$ \\
			\hline

	\end{tabular}}

	\caption{EER  performance indicator of multi snapshot fusion,  for the second scenario of the first protocol with $90\%$ confidence interval, where, Method is the feature extraction technique;
	Indicator is biometric performance parameter (EER) given in \%; Data set band is the wavelength used to collect each data set. }
	
	\label{fusion first protocol second scenario}
	
\end{table*}

\paragraph{Result on protocol-2} The same principle of Section.~\ref{Result on protocol-2} followed for performance evaluation. The unique scenario of this protocol measures the performance as function of sessions variations. For each subject, the templates and probs sets are extended with  evenly  size equal to $15$.  Consequently, the obtained genuine scores are $370 \times 15 = 5700$ and for imposter $380 \times 379 \times 15 = 2160300$ scores used for comparison. Table.~\ref{fusion second protocol} below reports the performance evaluation  of the proposed fusion scheme and recent feature extraction methods. Among all techniques, BSIF exhibits the best performance with $EER= 4.82 \pm 0.21\%$, $GAR @ EER= 95.18 \pm 0.22\%$, $HTER= 4.31 \pm 0.05\%$, and $Rank-1 = 90.75 \pm 0.89\%$  when data set band equal to $0960nm$. As well as, the least efficient results appears when $LTP$ used with data set band $0700nm$, where  $EER= 28.37 \pm 0.22\%$, $GAR @ EER= 71.63 \pm 0.23\%$, $HTER= 27.65 \pm 0.27\%$, and $Rank-1 = 23.18 \pm 2.41\%$. \\

Our results confirm what many previous works suggested when using more than one sample both from the template and prob side \cite{7 introduction,8 introduction,fusion2,fusion3}; although one single impression is similar to another one, their individual features sets contains complementarity information. The average operation we performed on such feature set allow reducing, as a low-pass filter, the eventual "noise" due to artifacts and boosting strongly the individual information which makes palmvein texture unique from person to person.

	\begin{table*}[hbt!]
	\centering
	\resizebox{0.8\textwidth}{!}{
		
		\begin{tabular}{@{}|c|c|c|c|c|c|@{}} \hline \hline
			
			\multicolumn{5}{|c|}{ \cellcolor {LightCyan} $\mathbf{Second\, protocol \ ( 15 \  templates \ \  \& \ \ 15\ probs )}$} \\
			\hline
			
			\multirow{2}{*} {$\mathbf{Method}$} & \multirow{2}{*} {$\mathbf{Indicator}$} & \multicolumn{3}{c|} {$\mathbf{Data \, set\, band}$} \\
			
			\cline{3-5} & & {$\mathbf{0700\textit{nm}}$} & {$\mathbf{0860\textit{nm}}$} & {$\mathbf{0960\textit{nm}}$} \\
			\hline

			\multirow{4}{*}{$\mathbf{LTP}$} &  {${EER}$} &$28.37 \pm 0.22$& $22.29\pm0.08$&$23.43\pm0.06$\\ \cline{2-5} 
			& 	${GAR @ EER}$ & $71.63\pm0.23$ & $77.72\pm0.12$ & $76.59\pm0.04$ \\ \cline{2-5}
			
			&{${minHTER}$} & $27.65\pm0.27$ & $21.57\pm0.10$ & $22.81\pm0.22$ \\ \cline{2-5}
			&	{${Rank-1}$} & $ 23.18\pm2.48$ & $33.38\pm0.74 $ & $33.96\pm0.02$\\
			\hline
			
			\multirow{4}{*}{$\mathbf{LDP}$} &  {$EER$} & $13.95\pm0.28$ & $12.47\pm0.09$ & $13.73\pm0.26$ \\ \cline{2-5} 
			& 	{${GAR @ EER}$} &$86.06\pm0.25 $ &$ 87.54\pm0.09 $& $86.26\pm0.23$ \\ \cline{2-5}
			&{	${minHTER}$} & $13.89\pm0.28$ &$12.16\pm0.18$ & $13.59\pm0.21$ \\ \cline{2-5}
			&	{${Rank-1}$} & $33.91\pm2.25$ & $41.32\pm1.57$& $38.79\pm0.81$\\
			\hline

			\multirow{4}{*}{$\mathbf{LPQ}$} &  ${EER}$ &$ 13.03\pm0.12 $& $18.08\pm0.21$ & $14.22\pm0.02$  \\ \cline{2-5} 
			& 	{${GAR @ EER}$} &$86.96\pm0.14 $ &$ 81.95\pm0.20$ & $85.77\pm0.03$ \\ \cline{2-5}
			&	{${minHTER}$} & $12.96\pm0.09$ &$18.00\pm0.17 $ & $13.95\pm0.19$ \\ \cline{2-5}
			&	{${Rank-1}$} & $45.85\pm2.24$ &$32.27\pm1.46$ &$ 48.32\pm0.42$\\
			\hline
			
			\multirow{4}{*}{$\mathbf{LBP}$} &  {$EER$} & $12.34\pm0.12$ & $12.58\pm0.27$ & $11.90\pm0.11$ \\ \cline{2-5} 
			& 	{${GAR @ EER}$} & $87.66\pm0.13$ & $ 87.43\pm0.27$& $88.09\pm0.12$ \\ \cline{2-5}
			&	{${minHTER}$} & $ 12.28\pm0.12$& $12.41\pm0.44$ & $11.86\pm0.08 $\\ \cline{2-5}
			&	{${Rank-1}$} & $49.28\pm2.42$ & $46.43\pm3.28$ & $53.11\pm1.11$\\
			\hline
			
			\multirow{4}{*}{$\mathbf{BSIF}$} &  {$EER$} & $7.64\pm0.01$ &  $11.01\pm0.24$& $\mathbf{4.82\pm0.21}$ \\ \cline{2-5} 
			& ${GAR @ EER}$&  $92.35\pm0.03$ & $89.00\pm0.23$ &  $\mathbf{95.18\pm0.22 }$ \\ \cline{2-5}
			&	${minHTER}$ & $7.40\pm0.05$ & $10.30\pm0.01$ &  $\mathbf{4.31\pm0.05 }$\\ \cline{2-5}
			&	${Rank-1}$ & $82.36\pm0.74$ & $71.24\pm4.84$ & $\mathbf{90.75\pm 0.89}$\\
			\hline
			
	\end{tabular}}
	
	\caption{Verification and identification performance of multi snapshot fusion.  The second protocol with $90\%$ confidence interval is considered, where, Method is the feature extraction technique; Indicator is biometric performance parameter given in \%; Data set band is the wavelength used to collect each data set.  }
	
	\label{fusion second protocol}
	
\end{table*}

			\subsection{Comparison with previous works}
				
				Palmvein recognition based on multi-snapshot fusion at feature level  is proposed in this work, furthermore, five of the state of the art textural feature (LTP, LDP, LBP, LPQ, and BSIF) extraction methods used. LBP and LPQ, and other textural descriptors are proposed in the context of palmvein recognition \cite{2 LBP section 2}, however, little works were presented based on BSIF features. The uniqueness of our proposed wok is the employment of BSIF  which not fully investigated for palmvein; several filters and patch sizes are offered with BSIF to show its preference for textural feature extraction, which helped to propose multi-snapshot fusion with different precisions to generalize the recognition stage and outperforming the over-fitting issue. Other uniqueness factor of our proposed work, is the performance evaluation which reported based on the statistical significance that hasn't completely shown in the context of palmvein to show the robustness and to facilitate  repeat the experimental works by any forthcoming works, thus motivate us to propose our work with respect to ISO terminology and different performance indicators which not observed together in one work. The following description shows the differences between our proposed work and the other recent works, see Table.~\ref{comparsion to other work}.    
				
				Zhou \textit {et al.} \cite{1 introduction} used several representations in the term of identification,  they used Hessian Phase, LRT, Ordinal Code, and Laplacianpalm, with identification rate at rank-1: $95\%$, $96\%$, $97\%$, and $77\%$ respectively, but the highest identification rate at rank-1 was $98\%$ and obtained from combinations of these algorithms. Although Zhou \textit {et al.}  achieved $0.28\%$ their results wasn't much conceiving, because they employed only 100 subjects to evaluate the performance and the best result requires to combine four different algorithms leading to increase the computational burden and  time consuming.  LBP and MF-LDTP were used in \cite{final} where the identification rates at rank-1 are equal to  $92.5\%$ and $95\%$, for LBP for MF-LDTP respectively, moreover the best obtained $EER$ was equal to $5\%$. We  compare our  results with that reported in Table~\ref{comparsion to other work}, and related to several recent works regarding to palmvein verification. It is clearly to conclude that the performance of our approach is at the best level of the state of the art, especially, when referring to multi-snapshot fusion, which seems to exploit the complementary information in multiple impressions of the same palmvein at best. 
				
				Our results built based on Raghavendra's \textit{et al.} \cite{ragavendar} protocols, and they used wavelets features, CASIA data set, and score level fusion. They  obtained the best $EER=1.64 \pm 0.45$ by combining the all weighted scores of all data set bands, meaning that during system operation, there is a need to have multi-capturing sensor with multi bands which increase the complexity of the system and reduce the user acceptability for biometric technology. On comparing our results with Raghavendra's \textit{et al.} results, we obtained $EER=1.74 \pm 0.87$ where feature level fusion is considered with single band of capturing sensor, also we compared more scores of genuine and impostor than them.

				\begin{table*}[hbt!]
					\resizebox{0.99\textwidth}{!}{
						\centering
						
						\begin{tabular}{@{}|c|c|c|c|c|c|@{}} \hline \hline
							
							\cellcolor {LightCyan} \textbf {Ref.}  & {EER} & {Comparator} & {Database} & {Fusion} & {Year}\\
							\hline
							\rowcolor{lightgray}
							\cite{table2 3} & Not mentioned  & Template matching & PolyU  & Not applied & 2007  \\
							\hline
							
							\cite{1 introduction} & 2.80\% & Key-point Comparator  & CASIA & Applied & 2010 \\
							\hline
							
							\rowcolor{lightgray}	

							\cite{table2 1} & 0.72\%  & Angular distance   & Private & Applied & 2011  \\
							\hline

							\cite{6 introduction} & Not mentioned & Hamming distance  & Private & Not applied & 2012 \\
							\hline
							
							\rowcolor{lightgray}
							
							\cite{ragavendar} & 1.64\% & Sparse representation & CASSIA & Applied & 2013 \\
							\hline

							\cite{PolyU1} &  3.60\%  & Cosine similarity & PolyU & Not applied & 2014 \\
							\hline

							\rowcolor{lightgray}
							\cite{table2 2} &  Not mentioned  &  Template matching  & Private  & Applied &  2014 \\
							\hline

							\cite{2 LBP section 2}  & 2.67\% &  Histogram matching & CASIA & Applied & 2014 \\
							\hline	
							 
							\rowcolor{lightgray} 
							 \cite{motivation}  & 0.1.647\% & Euclidean distance & PolyU multi- spectral & Applied  &  2014 \\
							 \hline

							\cite{ltp} & 0.00\% &  Cosine similarity  & PUT vein & Applied & 2015 \\ 
							\hline
							
							\rowcolor{lightgray}
							\cite{7 introduction}  &  1.60 &  Euclidean distance  &  CASIA & Applied & 2015 \\
							\hline
							\cite{final} & 5.00\% & Chi-square & CASIA &  Not applied & 2015 \\
							\hline
							
							\rowcolor{lightgray} 
							{Our proposed method}  & $1.64 \pm 0.45\%$ & Knn with Euclidean distance & PolyU hyper- spectral & Applied  &  2019 \\
							\hline

					\end{tabular}}
					
					\caption{Performance of some recent  hand crafted based works. Ref.: the related reference in the bibliography. Comparator: the matching system used in each work. Database: related data set used in each work.}
					
					\label{comparsion to other work}
					
				\end{table*}
				
				Among others, Elnasri \textit{et al.} \cite{PolyU1} proposed palmvein verification based on liner discrimination analysis depending on PolyU database. They claimed that the best $EER$ was equal to $0.0\%$ by considering six templates and 4 probes (several templates and probes were considered in that work with several results) for each subject but the performance decreased when probs increased. Moreover, they didn't used the whole data set in performance evaluation; they only used $2000$ images taken into account $10$ images for each subject. Moreover, Al-juboori \textit{et al.} \cite{motivation} used PolyU multispectral palmprint data set which is different than our data set,  and they utilized it in the context of palmvein recognition by halfing the dtat set for training and testing stages, \textit{i.e.}, 6 templates and 6 probs for each subject. Moreover, they considered multi- algorithmic fusion at the score level  which is considered time consuming with claimed $EER = 0.1378\%$. However, We employed three data sets with different bands and compared scores of 28 probs (utilized for testing) and 6 templates (utilized  for training) for each subject towards better generalization, with $EER=1.74 \pm 0.87\%$ in fusion scheme considering the over-fitting and bias estimation issues.
				
			Local texture patterns were proposed in \cite{LBP section 2}, LBP and LDP considered with $miniHTER= 6.67\%$ and $3.24\%$, respectively. Our proposed work reports that depending on BSIF and no fusion case the $miniHTER= 2.83 \pm 0.91\%$, also by considering fusion we got $miniHTER= 1.50 \pm 0.71\%$. In \cite{ltp}  the reported $EER$ is equal to $0$ but the feature vector was $1 \times 28400,$ that means more computational time required for templates generation, especially if compared with our proposed feature set size which is $1 \times 256$: for example, the BSIF algorithm employs less than $50 msec$ to generate a template (to extract features and represent the data based on histogram representations) on our platform based on Matlab $7.9.0 \ r2009b$, Windows $7$ PRO  $64$ bit, Pentium(R) Dual-Core CPU ES200 $  2.50 \ GHz$,   8 GB RAM.  
		
Finally, it can be pointed from Table \ref{comparsion to other work} that where private data sets are used, the performance can be referred only to those data sets. Standardized data size in training and performance sets should be adopted in order to have fully comparable experimental and reproducible results.

				\section{Conclusions}
				
				In this paper, we proposed a palmvein recognition for both tasks, verification and identification based on the BSIF algorithm for feature extraction and several feature-level fusion methods applied for sake of comparison. Our goal was to show the effectiveness of BSIF for palmvein biometric and how the verification and identification errors are reduced by fusion methods. On the basis of reported results, BSIF was not only the best method among others, but it greatly helped improving the performance when fusion is considered. The best results obtained by multiple snapshots confirmed that more information is extracted from different images using BSIF alone. Future works will be focused on a more extensive experimental investigation, but also to better understand the origin of BSIF effectiveness and ability to extract complementary features from multiple samples of the same image. Finally, the forthcoming works  must be taken in account   the user acceptability, computational burden, and over-fitting when considering fusion, since it is noticed that a lot of recent works missing to fix some of these issues.

			\end{document}